\documentclass[letterpaper, 10 pt, conference]{ieeeconf}  

\IEEEoverridecommandlockouts                              

\usepackage{array}
\usepackage[dvipsnames]{xcolor}
\usepackage{graphics} 
\usepackage{epsfig} 
\usepackage{mathptmx} 
\usepackage{times} 
\usepackage{amsmath} 
\usepackage{amssymb}  
\usepackage[citestyle=numeric-comp, sorting=none]{biblatex}
\usepackage{color, soul}
\usepackage{csquotes}
\usepackage{algorithm, algpseudocode}
\usepackage{url} 

\bibliography{references}

\title{\LARGE \bf
Dynamic Neural Potential Field: Online Trajectory Optimization in the Presence of Moving Obstacles
}

\author{
Aleksei Staroverov$^{1,2,3}$, Muhammad Alhaddad$^{3}$, Aditya Narendra$^{3}$\\
Konstantin Mironov$^{4}$, and Aleksandr Panov$^{1,3}$
\thanks{$^{1}$Cognitive AI Systems Lab}%
\thanks{$^{2}$National University of Science and Technology MISIS}%
\thanks{$^{3}$Moscow Independent Research Institute of Artificial Intelligence}%
\thanks{$^{4}$Ufa University of Science and Technology}%
}

\begin{document}

\maketitle
\thispagestyle{empty}
\pagestyle{empty}


\begin{abstract}

Generalist robot policies must operate safely and reliably in everyday human environments such as homes, offices, and warehouses, where people and objects move unpredictably. We present Dynamic Neural Potential Field (NPField-GPT), a learning-enhanced model predictive control (MPC) framework that couples classical optimization with a Transformer-based predictor of footprint-aware repulsive potentials. Given an occupancy sub-map, robot footprint, and optional dynamic-obstacle cues, our NPField-GPT model forecasts a horizon of differentiable potentials that are injected into a sequential quadratic MPC program via L4CasADi, yielding real-time, constraint-aware trajectory optimization. We additionally study two baselines: NPField-StaticMLP, where a dynamic scene is treated as a sequence of static maps; and NPField-DynamicMLP, which predicts the future potential sequence in parallel with an MLP.

In dynamic indoor scenarios from BenchMR and on a Husky UGV in office corridors, NPField-GPT produces more efficient and safer trajectories under motion changes, while StaticMLP/DynamicMLP offer lower latency. We also compare with the CIAO* and MPPI baselines. Across methods, the Transformer+MPC synergy preserves the transparency and stability of model-based planning while learning only the part that benefits from data: spatiotemporal collision risk. Code and trained models are available at \url{https://github.com/CognitiveAISystems/Dynamic-Neural-Potential-Field}.
    
\end{abstract}

\section{Introduction}
\begin{figure*}[htbp]
\centering
\includegraphics[width=0.95\textwidth]{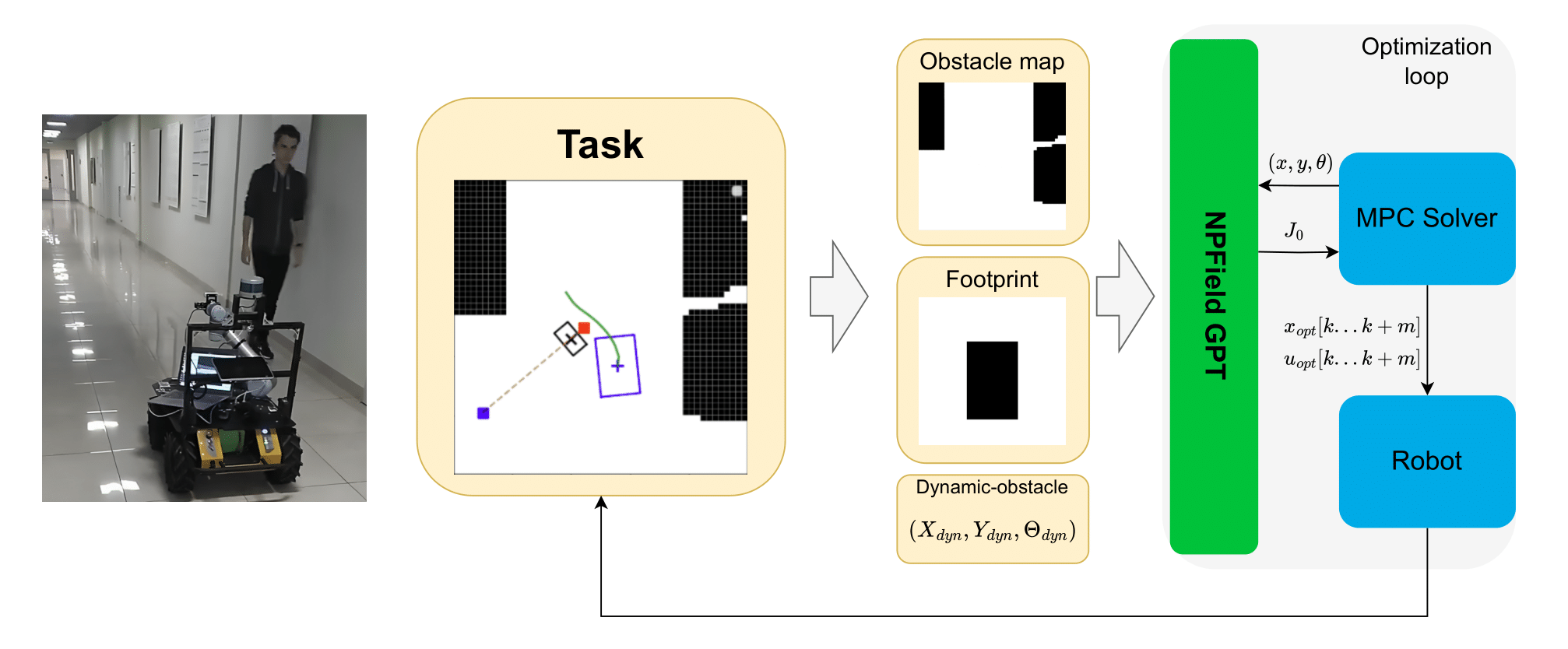}
\caption{Overview of NPField-GPT integrated with MPC for dynamic obstacle avoidance. Inputs include occupancy map, robot footprint, robot pose, and dynamic-obstacle state $\left(x_{dyn},y_{dyn},\theta_{dyn}\right)$. The neural model predicts a footprint-aware repulsive potential horizon, which is consumed by MPC to optimize $\mathbf{x}_{opt}$ and $\mathbf{u}_{opt}$ in real time.}
\label{fig:vis_abstract}
\end{figure*}

Mobile robots that act autonomously within human-oriented environments can become significant assistants for humans. They can be applied in offices, shops, homes, and medical facilities. Autonomous operation requires methods for planning motion within the environment. Trajectory planning is often executed in two stages: first, a rough global path is generated via search-based \cite{har68, nas07} or sampling-based \cite{kav96, lav01} methods; second, the global path is turned into a local trajectory under kinodynamic constraints and obstacle avoidance. The second stage is often executed online with a receding horizon strategy. Solutions for receding-horizon trajectory planning may be obtained with model predictive path integral (MPPI) \cite{wil16, wil17} or numerical model predictive control (MPC) \cite{bla11, ji16, sch20a, sch20b, zuo20, boj21, zen21, thi22}. MPPI can work with arbitrary obstacle maps, but it can produce invalid or chattering solutions. Numerical MPC provides fast and stable solutions; however, it requires an analytical representation of collision danger either as a set of constraints \cite{sch20a} or as a cost term, the repulsive potential. The term ``repulsive potential'' was introduced for the artificial potential field global planner \cite{kha85}. In MPC, the trajectory is biased toward safer regions according to the repulsive potential. In this work, we consider local planning with numerical MPC in the presence of dynamic obstacles (Fig. \ref{fig:vis_abstract}).

Avoiding collisions with dynamic obstacles (e.g., walking human beings in an office environment) is especially challenging. It means that the obstacle map for each frame within the prediction horizon will be different. The prediction of obstacle flow is a distinct task \cite{jai20, sha22, mur24}, which is out of our scope. It can be solved with various methods. Here, we consider dynamic obstacles to be predictable. The dynamic appearance of the obstacle model enlarges the computational complexity of the MPC problem due to a higher number of obstacle parameters.

We follow a hybrid design principle: learn only spatiotemporal collision risk and keep optimization model-based. Specifically, we extend NPField to dynamic scenes with three neural variants that produce differentiable, footprint-aware obstacle potentials for MPC.

Our framework comprises three architectures to balance accuracy and latency. The simplest variant, StaticMLP, processes dynamic environments as sequential static frames. To improve temporal reasoning, DynamicMLP infers future potentials concurrently using parallel heads. Finally, NPField-GPT employs a non-autoregressive Transformer to predict the entire potential horizon holistically, significantly enhancing the representation of spatial and temporal obstacle dynamics. In the GPT model, coordinate information is emphasized through explicit dynamic-obstacle channels, relative-coordinate fusion, and near-obstacle weighted training.

Our main contributions are:
\begin{itemize}
    \item A real-time MPC-compatible dynamic neural potential framework with three variants (StaticMLP, DynamicMLP, GPT) covering accuracy-latency trade-offs.
    \item A novel NPField-GPT architecture that predicts horizon potentials in parallel and strengthens obstacle-relative geometry conditioning.
    \item Extensive comparisons with CIAO* and MPPI in BenchMR scenarios and real Husky UGV office-corridor experiments.
    \item Open-source implementation with L4CasADi integration for differentiable neural costs inside MPC.
\end{itemize}

\section{Related works}
This section discusses existing approaches to motion planning, especially local planning in the presence of moving obstacles and collision avoidance using neural networks. The general planning task consists of finding a trajectory from a given start to a given goal. Well-known global planners such as A* \cite{har68}, Theta* \cite{nas07}, PRM \cite{kav96}, or RRT \cite{lav01} generate a reference of intermediate positions of the robot (a global geometric path). Following this reference leads to approaching the destination point.
However, in the common case, the geometric path is unaware of how to provide smooth and collision-free motion between intermediate positions. Basic global planners have special extensions, which provide planning with respect to kinodynamic constraints. A* search may be executed on a lattice of kinodynamically feasible motion primitives \cite{but14}, which results in an executable trajectory instead of a rough geometric path. Motion primitives allow the planner to check that there are no collisions between the intermediate state, including collisions with dynamic obstacles \cite{phi11, lin21, yak22, ali23}. Sampling-based planners may also be extended to satisfy kinodynamic constraints \cite{pal14} and replan while avoiding dynamic obstacles \cite{ott16}. 

The aforementioned methods aim to add dynamic collision avoidance into the global planning procedure, which might be computationally excessive in case of long global plans and short local horizons. Prediction of dynamic obstacles is often obtained from actual sensor data and requires replanning within a relatively short prediction horizon. Therefore, we further narrow down to receding horizon approaches for local planning.

\subsection{Receding horizon planning and dynamic obstacles}
The task of local planning is to turn a fixed part of the rough global plan into a smooth trajectory segment under consideration of obstacles and kinodynamic constraints. Model Predictive Path Integral (MPPI) \cite{wil16} achieves this via sampling random trajectory segments and generating a good solution based on these samples. Collision check for sample trajectories may be systematically extended to the case of dynamic obstacles \cite{moh22, pat24}. Computational heaviness and nondeterministic nature are disadvantages of MPPI; therefore, it may be used in cases when the process model is too complicated for MPC. 
Numerical MPC solves the local planning task as an optimal control problem. Obstacle avoidance is formalized as a set of constraints (e.g. \cite{sch20a}) or as an additional cost term of the optimization problem. This formalization leads to additional problem parameters describing obstacle properties. The number of parameters should be small to preserve MPC performance. Many approaches approximate the obstacle map with a low-dimensional model. Either obstacles or free space may be approximated with a set of simple geometric figures: points \cite{ji16}, circles \cite{sch20a, zen21}, rectangles \cite{sch20b}, polylines \cite{zie14}, or polygons \cite{bla11, fra15, thi22}. Some of these works explicitly consider dynamic environments \cite{zie14, fra15}; others can be adapted straightforwardly (e.g., CIAO* \cite{sch20a, sch20b} uses independent approximations of the free space at each timestep; these approximations can reflect dynamic obstacles).
The limitations of geometric approximations are computational challenges and loss of fidelity. Many works (e.g. \cite{bla11, thi22}) do not consider how to obtain a geometric approximation from an arbitrary obstacle map. In practice, maps are often provided as occupancy grids (a matrix projected onto the map; zero values correspond to free-space cells, while ones correspond to occupied cells). High-resolution grids have too many parameters to be passed to the MPC solver in raw form. In the next subsection, we consider approaches where obstacle data are modeled with neural networks. 

\subsection{Trajectory optimization with neural collision model}
Learning collision model for numeric trajectory optimization was considered in several works \cite{ada22, sal23a, kur22, kat23, jac24}. It is challenging to balance the precision and computational complexity of the collision model; therefore, existing works are constrained in terms of computation time and complexity of the maps. \cite{ada22, sal23a, kur22, kat23} use neural models inspired by Neural Radiance Field \cite{mil20}. These models learn the structure of a single obstacle map (in 2D or 3D) and may be used for navigation within the learned map. \cite{ada22, sal23a} exploit network inference for trajectory optimization, while \cite{kur22, kat23} optimize trajectory within the learning procedure.
Instead of learning a single obstacle map, the models from \cite{npfield,jac24} take an obstacle representation as input. Both architectures include two submodels: the first reduces the dimensionality of the obstacle representation, while the second calculates the collision score. Therefore, the first submodel provides a compact vector of parameters for a real-time MPC solver, while the second is integrated into this solver using the L4CasADi \cite{sal23a} framework. The difference is that NPField \cite{npfield} utilizes occupancy grids for footprint-aware collision avoidance of a wheeled mobile robot, while \cite{jac24} utilizes depth images for 3D collision avoidance of an aerial robot.
To our knowledge, there are no neural obstacle models that provide MPC avoidance from dynamic obstacles in real time. \cite{kat23} exploits a NeRF learning procedure for trajectory optimization in the presence of moving obstacles; however, the computational procedure takes tens of seconds, which is non-real-time. This work aims to provide a model that allows online MPC collision avoidance with a neural model of static and dynamic obstacles.

\subsection{Positioning and novelty.} Unlike classical geometric MPC \cite{sch20a, sch20b} or sampling-based methods \cite{wil16}, our approach naturally encodes footprint-aware costs from raw occupancy grids. Furthermore, in contrast to prior neural models that learn single scenes offline \cite{ada22, kur22, kat23} or use non-map inputs \cite{jac24}, our framework provides real-time, differentiable repulsive potentials for dynamic obstacle avoidance directly within the MPC loop.

\section{Background}
Following NPField, we define the statement and notations for model predictive local planning. We consider a non-holonomic wheeled robot with a differential drive. System state vector $\mathbf{x}=\{x,y,\theta,v\}$ includes 2D robot coordinates $x$ and $y$, its orientation $\theta$ and linear velocity $v$ (directed according to $\theta$). Control vector $\mathbf{u}=\{a,\omega\}$ includes robot acceleration $a$ (directed according to $\theta$) and angular velocity $\omega$. We consider the model with continuous time dynamics and discrete time control (i.e., $u$ is considered to be constant within the single timestep). A formal statement for the trajectory optimization problem is the following:

\begin{subequations} \label{mpc_d}
\begin{equation} \label{mpc_d_1}
\begin{aligned}
\{\mathbf{x}_{opt}[i],\mathbf{u}_{opt}[i]\}_{i=k}^{k+m}=\\
\arg\min\sum_{i=k}^{k+m}( \lVert{\mathbf{x}[i] - \mathbf{x}_{r}[i]\rVert}_{w_{x}} + \lVert{\mathbf{u}[i]\rVert}_{w_{u}} + J_o(\mathbf{x}[i],\mathbf{p}_o[i]) ),
\end{aligned}
\end{equation}
s.t.
\begin{equation}\label{diff_drive_model}
    \frac{dx}{dt} = v\cos\theta, \quad
    \frac{dy}{dt} = v\sin\theta, \quad
    \frac{dv}{dt} = a, \quad
    \frac{d\theta}{dt} = \omega.
\end{equation}
\end{subequations}
Here $\mathbf{x}_{r}$ is a reference path, $\mathbf{p}_o$ is a vector of obstacle parameters. The total cost function consists of three terms: path following term ($\lVert{\mathbf{x}[i] - \mathbf{x}_{r}[i]\rVert}_{w_{x}}$ is a weighted distance between actual trajectory and a reference path), control minimization term ($\lVert{\mathbf{u}[i]\rVert}_{w_{u}}$ is a weighted norm of the control input), and repulsive potential $J_o$. The aforementioned statement may be adapted to systems with other dynamic models (e.g., holonomic or car-like robots): this requires the change of $\mathbf{x}$, $\mathbf{u}$, and \eqref{diff_drive_model}, while the obstacle model will keep the same. Neural potential function is a neural network, which is trained to predict $J_o$ based on $\mathbf{x}$ and $\mathbf{p}_o$. In NPField $\mathbf{p}_o$ is an embedding of the obstacle map obtained from the neural encoder. Reference potential for each point is calculated based on the signed distance function (SDF) to the obstacle border. The reference potential for the whole robot is chosen as the maximum potential of the points within its footprint. The reference potential is non-differentiable (as is the SDF), and the maximum over footprint cells is computed algorithmically, so it cannot be used directly in the MPC loop. Instead, reference potential is used to generate a dataset for training the neural potential field. The difference between static and dynamic environments is that in the first case $\mathbf{p}_o$ is constant for the whole trajectory, while in the second case it depends on time.

\begin{figure*}[htbp]
\centering
\minipage{0.26\textwidth}
  \centering
  \includegraphics[width=\linewidth]{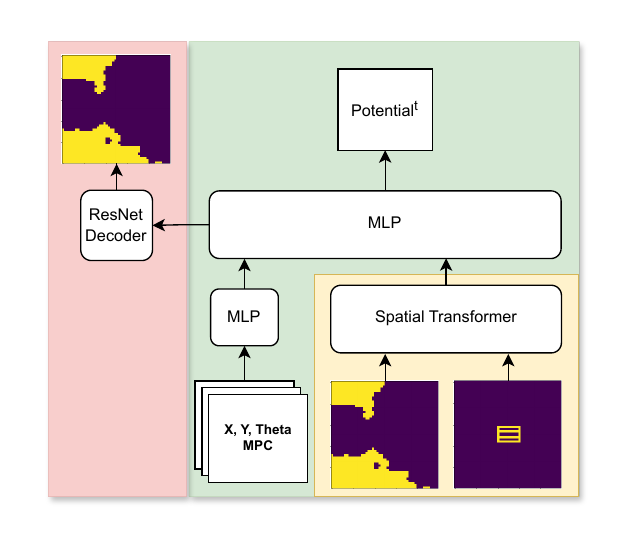}
\endminipage\hfill
\minipage{0.31\textwidth}
  \centering
  \includegraphics[width=\linewidth]{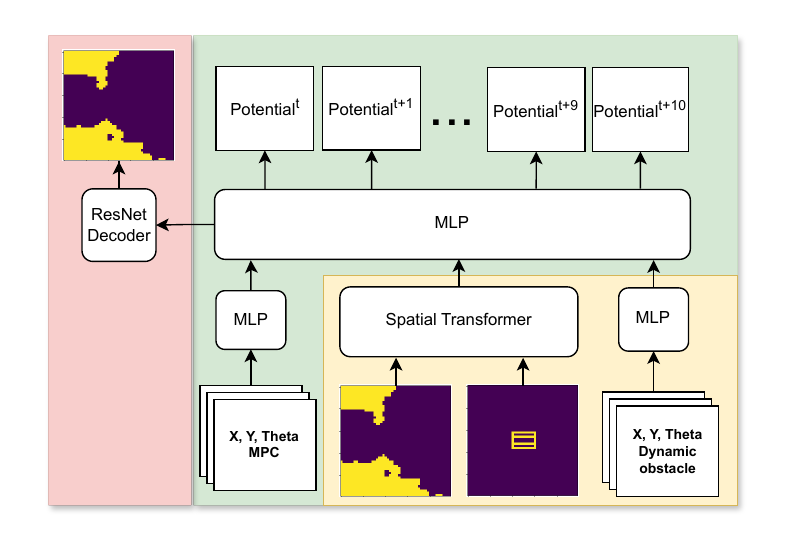}
\endminipage\hfill
\minipage{0.43\textwidth}
  \centering
  \includegraphics[width=\linewidth]{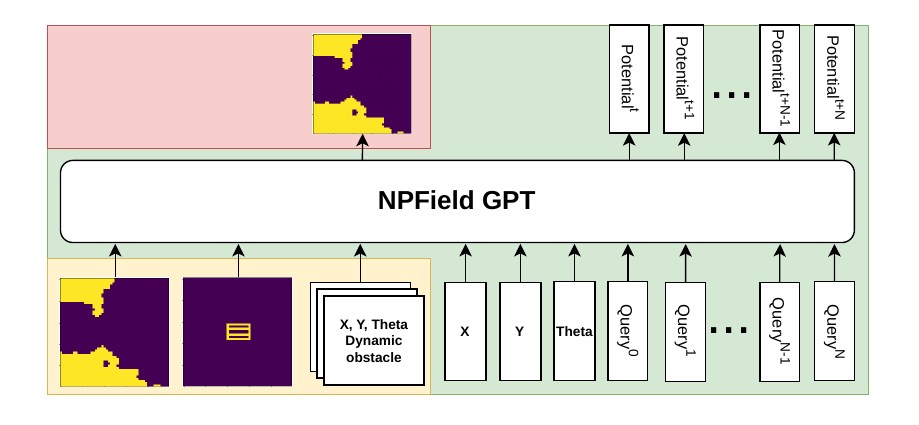}
\endminipage
\caption{StaticMLP (left), DynamicMLP (middle), and NPField-GPT (right). Yellow: map/footprint encoder; green: potential predictor for sampled states; red: auxiliary decoder used only during training. NPField-GPT predicts horizon potentials in one Transformer pass using learnable query tokens and coordinate-enhanced conditioning.}
\label{fig:net_arc_D1_D2_D3}
\end{figure*}

\paragraph{Why potentials instead of explicit trajectory prediction?}
Explicitly predicting obstacle trajectories can provide precise future states but introduces (i) an additional module to train and maintain, (ii) error compounding over the horizon, and (iii) coupling between prediction errors and MPC feasibility. Our approach estimates repulsive potentials that directly shape the MPC cost. This has two benefits: (a) optimization remains robust to small prediction errors because potential encodes spatial risk rather than a single hypothesized obstacle pose, and (b) potentials are differentiable and integrate naturally with L4CasADi for efficient gradient computation. When high-quality motion predictors are available, StaticMLP can consume predicted trajectories in the form of future obstacle maps, while DynamicMLP/NPField-GPT can be conditioned on predictor states $I_{dyn}=(x,y,\theta)$ and infer horizon risk directly from the current map context. Thus, our framework complements modern predictors by converting their outputs into optimization-friendly costs.


\section{Neural Network-Based Potential Field Generation}

We study three variants for producing footprint-aware repulsive potentials in dynamic scenes: NPField-StaticMLP, NPField-DynamicMLP, and NPField-GPT. All variants share a modular pipeline with three parts: (i) map/footprint encoding (yellow block), (ii) point-wise potential prediction for MPC query states (green block), and (iii) an auxiliary map-reconstruction decoder used only in training (red block). The auxiliary decoder regularizes the spatial embedding and is removed at inference time.

\textbf{NPField-StaticMLP} models dynamic navigation as a sequence of static maps and evaluates one-step potentials for each frame. This keeps the predictor simple but requires explicit future-map construction from obstacle forecasts.

\textbf{NPField-DynamicMLP} predicts all horizon steps in parallel with separate MLP heads conditioned on dynamic obstacle state $I_{dyn}=(x,y,\theta)$. This is efficient but temporal coupling between steps is weak because each step-specific head is independent.

\textbf{NPField-GPT} uses a Transformer backbone with non-autoregressive horizon prediction. Instead of generating step $t{+}1$ from step $t$, NPField-GPT appends learnable query tokens (10 for the horizon) and predicts the full potential sequence in a single forward pass.

\paragraph{Token sequence.} The input to the Transformer is a single sequence of tokens. The \emph{context} is built from 11 tokens in a fixed order: (1) map embedding (CNN plus spatial self-attention over the occupancy grid, projected to dimension $d$); (2) robot footprint embedding; (3)--(4) dynamic-obstacle position $(x_{dyn}, y_{dyn})$ each encoded by a linear layer; (5)--(6) dynamic-obstacle orientation $\theta_{dyn}$ as $\sin\theta_{dyn}$ and $\cos\theta_{dyn}$; (7)--(8) query robot position $(x, y)$; (9)--(10) query robot orientation $\theta$ as $\sin\theta$ and $\cos\theta$; (11) a learned \emph{relative-coordinate fusion} token that combines robot and obstacle coordinates via an MLP (context plus delta terms) and trainable gains to emphasize obstacle-relative geometry. The map and footprint encodings are computed once from the occupancy sub-map and footprint image; the coordinate tokens depend on the query pose $(x,y,\theta)$ and dynamic state $I_{dyn}=(x_{dyn},y_{dyn},\theta_{dyn})$. After the context, $H=10$ learnable query vectors are appended (the vectors themselves are fixed parameters, not a function of $(x,y,\theta)$). Learned positional embeddings are added to the full sequence; the stack is passed through the causal Transformer and layer norm. Because each query token attends over all preceding context tokens---including the robot pose $(x,y,\theta)$ and dynamic-obstacle state---the output at the query positions is conditioned on the query pose. The hidden states at the query positions are projected by a linear head and passed through sigmoid to obtain the predicted potential at each horizon step. Thus the potential horizon is predicted in one forward pass, with dependence on the query pose arising through attention to the context.

\paragraph{Training loss.} Let $\hat{y}_{b,h}$ and $y_{b,h}$ denote the predicted and target potential for batch index $b$ and horizon step $h$. Let $d_b$ be the Euclidean distance from the query pose $(x_b, y_b)$ to the dynamic obstacle $(x_{dyn,b}, y_{dyn,b})$. We use a distance-dependent weight $w(d) = 1 + \alpha \exp\bigl(-(d/\sigma)^2/2\bigr)$ with $\alpha=2$, $\sigma=0.5$ to upweight errors near the obstacle. The potential loss is the weighted mean squared error
\begin{equation}
\mathcal{L}_{\mathrm{pot}} = \frac{1}{B H} \sum_{b,h} w(d_b)\, (\hat{y}_{b,h} - y_{b,h})^2.
\end{equation}
An auxiliary occupancy reconstruction loss $\mathcal{L}_{\mathrm{map}}$ is computed by decoding the hidden state at the last context token (the relative-coordinate fusion token) through a small decoder and applying cross-entropy to the ground-truth occupancy grid; this stabilizes the spatial embedding. The total training loss is
\begin{equation}
\mathcal{L} = \mathcal{L}_{\mathrm{pot}} + \lambda_{\mathrm{map}}\, \mathcal{L}_{\mathrm{map}},
\end{equation}
with $\lambda_{\mathrm{map}} \geq 0$ (default 1). The auxiliary decoder and $\mathcal{L}_{\mathrm{map}}$ are used only during training and removed at inference.

\section{Dataset preparation and generation details}
\label{sec:dataset}
The dataset collection procedure includes the choice of static maps, robot footprints, and dynamic obstacles. Static maps were chosen based on the MovingAI \cite{stu12} city map dataset and an occupancy grid for an office building of the Institute. The Husky UGV footprint is used in this work.
In addition to the robot footprints, one footprint is used for the dynamic obstacle: a 0.7$\times$0.5\,m rectangle that moves in linear motion with a random orientation. Dataset collection includes the following steps:

\begin{figure}[htbp]
\centering
\includegraphics[width=0.4\textwidth]{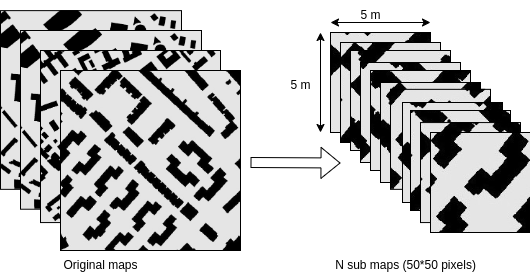}
\caption{Cropping original maps into sub-maps}
\label{fig:np_dataset1}
\end{figure} 

\begin{enumerate}
    \item The maps are cropped into sub-maps and represented as images with 50$\times$50 resolution (Fig.~\ref{fig:np_dataset1}), where each pixel corresponds to 0.1\,m (10$\times$10\,cm) in the real environment (i.e., each sub-map covers 5$\times$5\,m). This yields 1000 unique sub-maps for data generation.
    \item For each sub-map, one agent position with three orientations is sampled. Assuming 0.3\,m/s constant speed, 10 forward positions per orientation are generated (31 variants per sub-map: 1 static-only; 3$\times$10 with the agent).
   \item Every sub-map copy is transformed into a costmap where each cell is filled with the reference potential. \begin{enumerate}
    \item The signed distance function (SDF) is computed for each cell (distance to the nearest obstacle border; positive in free space, negative in obstacles). \item Repulsive potential per cell is $J_o = w_1(\pi/2 + \arctan(w_2 - w_2\,\mathrm{SDF}))$. This sigmoid is low far from obstacles, approaches $w_1$ asymptotically inside obstacles, and has maximum derivative on the obstacle border. 
    \end{enumerate}
    \item Two robot footprints are used, both corresponding to a real Husky UGV mobile manipulator (folded and outstretched arm).
    \item For each footprint, the potential at each query pose is obtained by placing the footprint at 10 random orientations per pixel for each copy of every sub-map; the robot footprint is projected onto the map and the maximum potential over the covered cells is taken as the repulsive potential. 
\end{enumerate}

The resulting dataset is tabular with fields: sub-map ID, query robot pose, dynamic obstacle pose/direction, footprint ID, and target potential.

Data splits and validation. We split the dataset into training/validation/test sets by sub-map (70/15/15\%). All figures that reference a \emph{validation} configuration use sub-maps and agent placements not present in training. We also include direction-reversal stress tests by flipping $I_{dyn}$ on held-out validation sub-maps to evaluate robustness.

\section{Implementation and computational profile}
\label{sec:implementation}

We solve the nonlinear MPC problem via Sequential Quadratic Programming (SQP) using Acados \cite{acados} with CasADi \cite{and19} for differentiation. To integrate our PyTorch \cite{pas19} potential model into the Acados solver, we wrap it using L4CasADi \cite{sal23a}, which provides an exact symbolic description suitable for accurate gradients (unlike local approximations such as ML-CasADi \cite{sal23}). This allows us to directly evaluate the predicted horizon potential in the Acados model, optimizing a nonlinear least-squares objective that combines tracking, control effort, and the neural potential term.

\paragraph{End-to-end runtime breakdown.} On an AMD Ryzen 5 3500 CPU and NVIDIA GeForce RTX 2080 (CUDA), typical per-replan times are:
\begin{itemize}
    \item map/footprint preprocessing and $h_{\text{map}}$ encoding (GPU): 6--12 ms;
    \item per-sample point embedding and NPField forward over the horizon (GPU): NPField-StaticMLP: 30--40 ms, NPField-DynamicMLP: 55--80 ms, NPField-GPT: 120--160 ms (Transformer with parallel query-token horizon prediction);
    \item Acados SQP solve with L4CasADi cost (CPU): 250--550 ms depending on variant and scenario complexity;
    \item ROS I/O and sensor updates: 10--20 ms.
\end{itemize}
This yields end-to-end medians consistent with Table~\ref{tab:comparative}. The bottleneck remains the MPC solve, while NPField-GPT adds moderate overhead due to the larger Transformer backbone and enhanced coordinate conditioning.

\paragraph{Scalability.} Let $N$ be the horizon length and $M$ the number of dynamic obstacles. For NPField-StaticMLP, complexity scales as $\mathcal{O}(N)$ forward calls of a single-step potential; for NPField-DynamicMLP, a constant number of heads gives $\mathcal{O}(1)$ forward depth per time step (batched over $N$); for NPField-GPT, horizon prediction is parallel in one Transformer pass, with complexity dominated by attention over context and query tokens. MPC cost is dominated by the nonlinear solve and scales roughly quadratically in $N$; we use warm starts and structured costs. With multiple obstacles, $I_{dyn}$ and token counts grow linearly in $M$; we cap $M$ per local window and prioritize nearest obstacles.

\paragraph{Training.} For NPField-GPT we use typical Transformer settings: 4 layers, 4 heads, embedding size 576; 10 learnable horizon query vectors; map encoder output projected to the model embedding width before Transformer fusion. Optimization uses an Adam-family optimizer with gradient clipping and optional mixed precision.

Our local planner works together with Theta* \cite{nas07} global planner, which generates global plans as polylines. Note that Theta* uses a simplified version of the robot footprint (a circle with a diameter equal to the robot width) as it fails to provide a safe path with a complete footprint model. 
This simplified model does not guarantee the safety of the global plan; therefore, the safety of the trajectory is provided by our local planner.

\section{Experiments}

\begin{figure}[htbp]
\centering
\includegraphics[width=.45\textwidth]{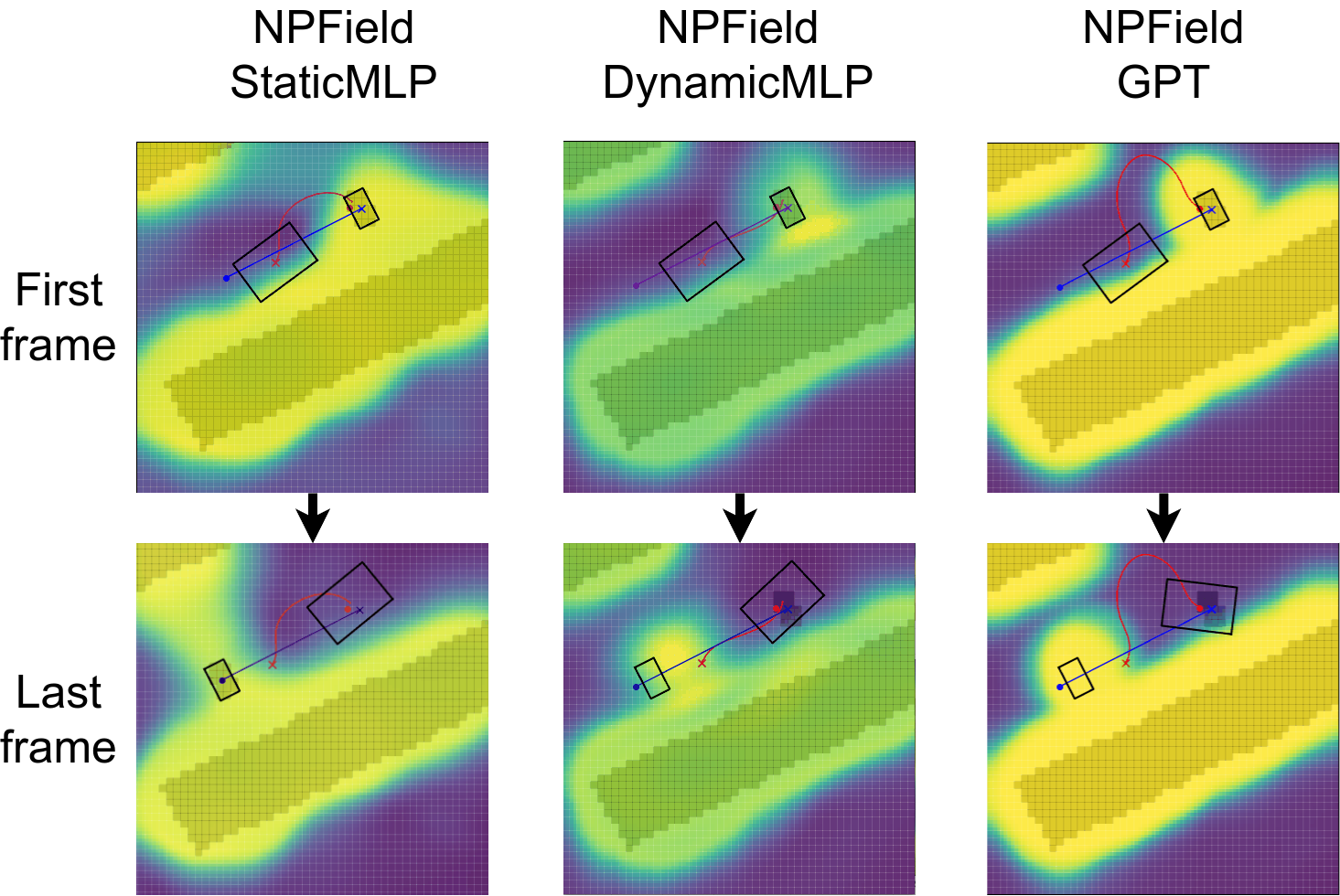}
\caption{First and last frames of a representative validation scenario for each method. The heatmap encodes the predicted repulsive potential. The larger rectangle denotes the robot footprint; the smaller rectangle denotes the dynamic obstacle.}
\label{fig:npfield_traj}
\end{figure}

\begin{figure*}[htbp]
\centering
\includegraphics[width=.90\textwidth]{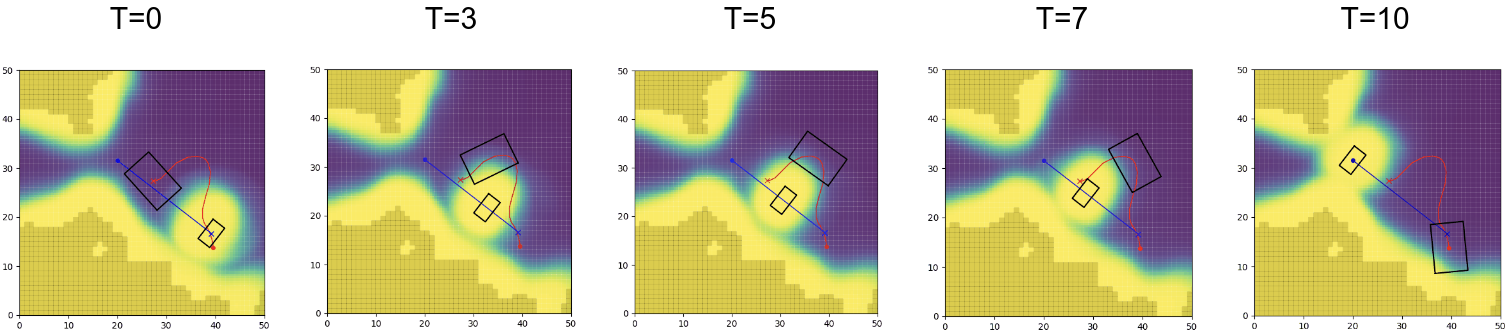}
\caption{Five sequential time steps of MPC trajectory optimization. The sequence illustrates how NPField-GPT adapts the full-horizon potential in real time, enabling smooth collision avoidance without explicit future-map rollout.}

\label{fig:traj_nav}
\end{figure*} 

\subsection{Numerical experiments}
We evaluated our algorithm on 100 scenarios using the \cite{heiden2021benchmr} framework, which includes tasks like navigating through narrow passages. Figure~\ref{fig:traj_nav} shows a representative example: five successive MPC replan steps in which NPField-GPT steers the robot around a moving obstacle by continuously updating the predicted potential horizon. The assessment covered standard metrics such as planning time, path length, smoothness, and angle-over-length, where a lower value is preferable for all. Additionally, we introduced a custom metric, ``safety distance'' (the minimum value of the SDF).

The experimental results, as detailed in Table~\ref{tab:comparative}, compared three versions of the Dynamic NPField algorithm: StaticMLP, DynamicMLP, and NPField-GPT, as well as MPPI and the CIAO* \cite{sch20a, sch20b} trajectory optimization algorithm. Figure~\ref{fig:npfield_traj} compares the predicted potential fields at the first and last frames of a validation scenario. The qualitative differences are clear: StaticMLP tends to merge the dynamic obstacle into the surrounding wall, losing its distinct boundary because the obstacle footprint is small relative to the map. DynamicMLP produces overly diffuse boundaries and does not always capture the temporal displacement of the obstacle correctly. NPField-GPT yields the sharpest and most accurate potentials, maintaining well-defined boundaries for both static walls and the moving obstacle throughout the horizon.

For each waypoint of the reference, CIAO* approximates the free space around the robot with a convex figure (circle or rectangle) and constrains the robot to be inside this figure. DynamicMLP and NPField-GPT directly use the data about the bounding box and motion direction of the dynamic obstacle, while StaticMLP and CIAO* require the projection of the predicted bounding boxes onto the obstacle map. 

Overall, CIAO* showed average performance, which aligns with its status as a state-of-the-art method. MPPI was faster than CIAO* but frequently failed to find optimal trajectories, resulting in generally inferior performance. StaticMLP was the fastest, registering a computation time of 366.62 ms, which is over 400 ms faster than CIAO*.
In contrast, NPField-GPT achieved the safest trajectory with the shortest path, as shown by the highest safety distance, lowest path length, and lowest AOL, though at the cost of increased computation time. This behavior suggests that NPField-GPT is the most appropriate choice for real-world scenarios where computation time can be traded for enhanced safety and path efficiency. In Table~\ref{tab:comparative}, we abbreviate planner names as NPF-SMLP, NPF-DMLP, and NPF-GPT.

\begin{table}[htbp]
\centering
\caption{Comparative studies on Dynamic obstacles scenarios}
\begin{tabular}{||m{1.3cm} |m{0.9cm}|m{0.8cm}| m{1.2cm} | m{0.7cm} | m{1cm} ||}
 \hline
  Planner & Time, ms  & Length, m & Smoothness & AOL & Safety distance, m \\ [0.5ex] 
 \hline\hline
 MPPI &  563.82 & 4.13 & 0.036  & 0.026  & 0.036 \\
 \hline
CIAO* & 780.83  & 3.62  & 0.028 & 0.019  & 0.112 \\
 \hline 
NPF-SMLP & \textbf{366.62} & 2.76 & 0.013 & 0.033  & 0.093 \\
 \hline
 NPF-DMLP & 493.50  & 2.15  & 0.017 & 0.058  & 0.109 \\
 \hline
 NPF-GPT & 663.81  & \textbf{2.06}  & \textbf{0.011} & \textbf{0.015} & \textbf{0.115} \\
 \hline
\end{tabular}
\label{tab:comparative}
\end{table}


\begin{figure}[htbp]
\centering
\includegraphics[width=0.45\textwidth]{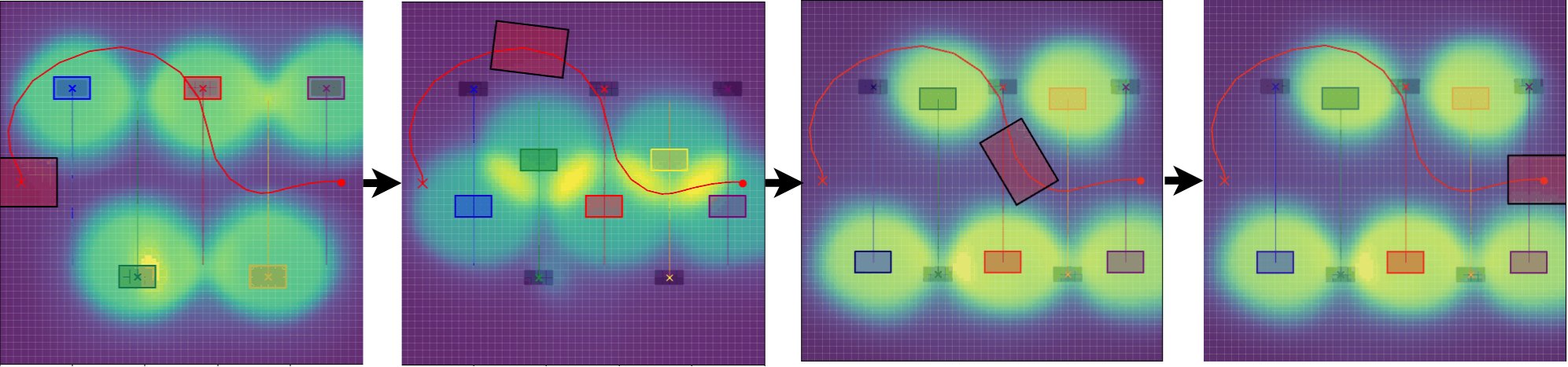}
\caption{Example of multi-obstacle scenarios solved by NPField-GPT. The potential field is obtained by summing per-obstacle predictions from the single-obstacle model without retraining.}
\label{fig:multiple_obstacles}
\end{figure}


\subsection{Extension to multiple dynamic obstacles}
NPField-GPT naturally extends to scenes with multiple dynamic obstacles via additive superposition of repulsive potentials, requiring no architectural changes or retraining. Given $M$ dynamic obstacles, the combined occupancy map (with all obstacles rendered at the current timestep) is encoded once, while the per-obstacle dynamic state $I_{dyn}^{(j)}=(x_{dyn}^{(j)},y_{dyn}^{(j)},\theta_{dyn}^{(j)})$ is supplied individually to produce $M$ independent embeddings. At each MPC query point, the model is evaluated $M$ times and the resulting horizon potentials are summed: $J_o^{\text{multi}} = \sum_{j=1}^{M} J_o^{(j)}$. This superposition is integrated into the Acados solver via a single L4CasADi wrapper that loops over embeddings internally. We evaluate this scheme on scenarios with 2--5 moving obstacles (Fig.~\ref{fig:multiple_obstacles}).

\subsection{Real robot experiments}
We tested the concept using a Husky UGV mobile manipulator as a ROS module. For local planning and control, we utilized MPC, and for global planning, we used the Theta* planner \cite{nas07}. Our testing scenario involved the robot maneuvering through a complex map. The entire navigation stack included a Cartographer \cite{hess2016real} and RTAB-Map \cite{rtabmap} for the global planner, with a second RTAB-Map for the local planner. All components of the control system were implemented as ROS nodes, with a central node managing communication with the Husky UGV hardware. Real-time planning and execution can be viewed in the supplementary video.

\section{Conclusion, limitations and future work}
This article presents a novel approach to MPC collision avoidance, which extends Neural Potential Field for the case of an environment with dynamic obstacles. We propose three neural architectures (NPField-StaticMLP, NPField-DynamicMLP, NPField-GPT) for predicting dynamic neural potentials and integrate them with real-time MPC.

Overall, the performance measures in Table~\ref{tab:comparative} show that our controllers require hundreds of milliseconds to replan the trajectory. This is sufficient for an indoor mobile robot with limited velocity and a 1--2 Hz replanning rate, while faster systems such as cars or drones require a higher rate. Performance may be improved by using more powerful hardware and developing faster models and implementations. 

It is worth noting that NPField is a learning-based method and therefore it has no theoretical safety guarantees. Therefore, the obtained solution has to be checked for collisions before its execution.

A significant assumption for the MPC problem is that for each dynamic obstacle, we know either a prediction of its future trajectory (NPField-StaticMLP) or a constant direction of its movement. We compensate for the inaccuracy of this assumption via replanning the trajectory on each step. Meaningful directions of future research include integrating neural potential forecasting with advanced motion prediction techniques, uncertainty-aware obstacle models, and joint multi-obstacle architectures that capture interaction effects beyond additive superposition.

\section*{Acknowledgments}

The study was supported by the Ministry of Economic Development of the Russian Federation (agreement No. 139-15-2025-013, dated June 20, 2025, IGK 000000C313925P4B0002).

\printbibliography

\end{document}